# Latent Structured Ranking

**Jason Weston**
Google, New York, USA.
jweston@google.com

**John Blitzer**
Google, Mountain View, USA.
blitzer@google.com

## Abstract

Many latent (factorized) models have been proposed for recommendation tasks like collaborative filtering and for ranking tasks like document or image retrieval and annotation. Common to all those methods is that during inference the items are scored independently by their similarity to the query in the latent embedding space. The structure of the ranked list (i.e. considering the set of items returned as a whole) is not taken into account. This can be a problem because the set of top predictions can be either too diverse (contain results that contradict each other) or are not diverse enough. In this paper we introduce a method for learning latent structured rankings that improves over existing methods by providing the right blend of predictions at the top of the ranked list. Particular emphasis is put on making this method scalable. Empirical results on large scale image annotation and music recommendation tasks show improvements over existing approaches.

## 1 INTRODUCTION

Traditional latent ranking models score the $i$th item $d_i \in \mathbb{R}^D$ given a query $q \in \mathbb{R}^D$ using the following scoring function:

$$f(q, d_i) = q^\top W d_i = q^\top U^\top V d_i, \qquad (1)$$

where $W = U^\top V$ has a low rank parameterization, and hence $q^\top U^\top$ can be thought of as the latent representation of the query and $V d_i$ is equivalently the latent representation for the item. The latent space is $n$-dimensional, where $n \ll D$, hence $U$ and $V$ are $n \times D$ dimensional matrices. This formulation covers a battery of different algorithms and applications.

For example, in the task of collaborative filtering, one is required to rank items according to their similarity to the user, and methods which learn latent representations of both users and items have proven very effective. In particular, Singular Value Decomposition (SVD) (Billsus and Pazzani, 1998; Bell *et al.*, 2009) and Non-negative Matrix Factorization (NMF) (Lee and Seung, 2001) are two standard methods that at inference time use equation (1), although the methods to learn the actual parameters $U$ and $V$ themselves are different. In the task of document retrieval, on the other hand, one is required to rank text documents given a text query. The classical method Latent Semantic Indexing (LSI) (Deerwester *et al.*, 1990) is an unsupervised approach that learns from documents only, but still has the form of equation (1) at test time. More recently, supervised methods have been proposed that learn the latent representation from (query, document) relevance pairs, e.g. the method Polynomial Semantic Indexing (SSI) (Bai *et al.*, 2009). Finally, for multiclass classification tasks, particularly when involving thousands of possible labels, latent models have also proven to be very useful, e.g. the WSABIE model achieves state-of-the-art results on large-scale image (Weston *et al.*, 2011) and music (Weston *et al.*, 2012) annotation tasks. Moreover, all these models not only perform well but are also efficient in terms of computation time and memory usage.

Scoring a single item as in equation (1) is not the end goal of the tasks described above. Typically for recomendation and retrieval tasks we are interested in ranking the items. This is achieved by, after scoring each individual item using $f(q, d_i)$, sorting the scores, largest first, to produce a ranked list. Further, typically only the top few results are then presented to the user, it is thus critical that the method used performs well for those items. However, one potential flaw in the models described above is that scoring items individually as in eq. (1) does not fully take into account the joint set of items at the top of the list (even when optimizing top-of-the-ranked-list type loss functions).

The central hypothesis of this paper is that latent ranking methods could be improved if one were to take into account the structure of the ranked list during inference. In particular this would allow the model to make sure there is the right amount of consistency and diversity in the predictions.

Let us suppose for a given query that some of the predictions at the top of the ranked list are accurate and some are inaccurate. A model that improves the consistency of the predictions might improve overall accuracy. A structured ranking model that predicts items dependent on both the query *and* other items at the top of the ranked list can achieve such a goal. To give a concrete example, in a music recommendation task you might not want to recommend both "heavy metal" and "60s folk" in the same top $k$ list. In that case, a structured model which encodes item-item similarities as well as query-item similarities could learn this by representing those two items with very different latent embedding vectors such that their pairwise item-item contribution is a large negative value, penalizing both items appearing in the top $k$. Note that a structured ranking model can do this despite the possibility that both items are a good match to the query, so an unstructured model would find this difficult to achieve.

Conversely, if improved results are gained from encouraging the top ranked items to be a rather diverse set for a particular query, then a structured model can learn to predict that instead. For example in the task of document retrieval, for ambiguous queries like "jaguar", which may refer either to a *Panthera* or to the car manufacturer, diversity should be encouraged. The goal of a structured ranker is to learn the optimal tradeoff between consistency and diversity on a case-by-case (per query) basis. As latent parameters are being learnt for each query type this is indeed possible.

In this work we propose a latent modeling algorithm that attempts to do exactly what we describe above. Our model learns to predict a ranked list that takes into account the structure of the top ranked items by learning query-item and item-item components. Inference then tries to find the maximally scoring set of documents. It should be noted that while there has been strong interest in building structured ranking models recently (Bakir *et al.*, 2007), to our knowledge this is the first approach of this type to do so for latent models. Further, the design of our algorithm is also particularly tuned to work on large scale datasets which are the common case for latent models, e.g. in collaborative filtering and large scale annotation and ranking tasks. We provide empirical results on two such large scale datasets, on a music recommendation task, and an image annotation task, that show our structured method brings accuracy improvements over the same method without structure as well as other standard baselines. We also provide some analysis of why we think this is happening.

The rest of the paper is as follows. Section 2 describes our method, Latent Structured Ranking (LaSR). Section 3 discusses previous work and connects them to our method. Section 4 describes our empirical results and finally Section 5 concludes.

## 2 METHOD

Given a query $q \in \mathcal{Q}$ our task is to rank a set of documents or items $\mathcal{D}$. That is, we are interested in outputting (and scoring) a permutation $\bar{d}$ of the set $\mathcal{D}$, where $\bar{d}_j$ is the $j$th item in the predicted ranked list. Our ultimate goal will be to design models which take into account not just individual document scores but the (learned) relative similarities of documents in different positions as well.

### 2.1 SCORING PERMUTATIONS BY SCORING INDIVIDUAL ITEMS

Let us begin by proposing methods for using the standard latent model of eq. (1) to score permutations. We need a method for transforming the scores for single documents into scores for permutations. Such transformations have been studied in several previous works, notably (Le and Smola, 2007). They show that finding maximally scoring permutations from single documents can be cast as a linear assignment problem, solvable in polynomial time with the Hungarian algorithm.

For the vanilla model we propose here, however, we can use a simple parameterization which allows for inference by sorting. For any given permutation we assign a score as follows:

$$f_{vanilla}(q, \bar{d}) = \sum_{i=1}^{|\bar{d}|} w_i (q^\top U^\top V \bar{d}_i), \quad (2)$$

where for each position $i$ in the permutation, we associate a weight $w_i$, where $w_i$ can be any weights such that $w_1 > w_2 > \cdots > w_{|\bar{d}|} \geq 0$. For example, one can just set $w_i = \frac{1}{i}$. Inference using this model is then performed by calculating:

$$F_{vanilla}(q) = \text{argmax}_{\bar{d}'}[f_{vanilla}(q, \bar{d}')].$$

In this case, computing the best-scoring assignment is simply a matter of sorting documents by their scores from eq. (1). To see this note that the score of any unsorted pair can be increased by sorting, since the positional weights $w_i$ are fixed and decreasing.

## 2.2 LATENT STRUCTURED RANKING

The fundamental hypothesis of this paper is that including knowledge about the structure of the rankings at inference time will improve the overall set of ranked items. That is, we want to define a model where the score of a document $\bar{d}_i$ does not only depend on the query $q$ but also on the other items and their respective positions as well. What is more, we would prefer a model that places more weight on the top items in a permutation (indeed, this is reflected by common ranking losses like MAP and precision@k).

This leads us to propose the following class of Latent Structured Ranking (LaSR) models:

$$f_{lsr}(q, \bar{d}) = \sum_{i=1}^{|\bar{d}|} w_i(q^\top U^\top V \bar{d}_i) + \sum_{i,j=1}^{|\bar{d}|} w_i w_j (\bar{d}_i^\top S^\top S \bar{d}_j) \,. \quad (3)$$

In addition to the parameters of eq. (2), we now introduce the additional parameter $S$. $S$ takes into account the structure of the predicted ranked list. $S^\top S$ is a low rank matrix of item-item similarities where $S$ is a $n \times D$ matrix, just like $U$ and $V$, and must also be learnt by the model using training data.

## 2.3 CHOICE OF THE $w_i$ PARAMETERS

The weights $w$ are crucial to the usefulness of the matrix in the second term of eq. (3). If $w_i = 1$ for all $i$ then the entire second term would always be the same no matter what choice of ranking $\bar{d}$ one chooses. If the position weights $w_i$ are decreasing, however, then the structural term $S$ is particularly meaningful at the top of the list.

As suggested before in Section 2.1 we could choose $w_i = \frac{1}{i}$. In that case the items that are at the top of the predicted ranked list dominate the overall score from the second term. In particular, the pairwise item-item similarities between items in the top-ranked positions play a role in the overall choice of the entire ranked list $\bar{d}$. Our model can hence learn the consistency vs. diversity tradeoff within the top $k$ we are interested in.

However, if one knows in advance the number of items one wishes to show to the user (i.e. the top $k$) then one could choose directly to only take into account those predictions:

$$w_i = 1/i, \text{ if } i \leq k, \text{ and 0 otherwise.} \quad (4)$$

As we will see this also has some computational advantages due to its sparsity, and will in fact be our method of choice in the algorithm we propose.

## 2.4 MODEL INFERENCE

At test time for a given query we need to compute:

$$F_{lsr}(q) = \text{argmax}_{\bar{d}'}[f_{lsr}(q, \bar{d}')]. \quad (5)$$

Just as inference in the vanilla model can be cast as a linear assignment problem, inference in the LaSR model can be cast as a quadratic assignment problem (Lacoste-Julien et al., 2006). This is known to be NP hard, so we must approximate it. In this section, we briefly discuss several alternatives.

- Linear programming relaxation: Since we know we can cast our problem as quadratic assignment, we could consider directly using the linear programming relaxation suggested by (Lacoste-Julien et al., 2006). In our experiments, however, we have tens of thousands of labels. Solving even this relaxed LP per query is computationally infeasible for problems of this size. We note that WSABIE's (Weston et al., 2011) sampling-based technique is an attempt to overcome even linear time inference for this problem.

- Greedy structured search: we could also consider a greedy approach to approximately optimizing eq. (5) as follows: (i) pick the document $\bar{d}_1 \in \mathcal{D}$ that maximizes:

$$f_{greedy}(q, \bar{d}_1) = w_1(qU^\top V \bar{d}_1) + (w_1)^2(\bar{d}_1^\top S^\top S \bar{d}_1) \quad (6)$$

and then fix that document as the top ranked prediction. (ii) Find the second best document dependent on the first, by maximizing (for $N = 2$):

$$f_{greedy}(q, \bar{d}_N) = w_N(qU^\top V \bar{d}_N)q$$
$$+ \sum_{i=1}^{N} w_i w_N (\bar{d}_i^\top S^\top S \bar{d}_N).$$

Finally, (iii) repeat the above, greedily adding one more document each iteration by considering the above equation for $N = 3, \ldots, k$ up to the number of desired items to be presented to the user.

This method has complexity $O(k^2|\mathcal{D}|)$. Its biggest drawback is that the highest-scoring document is chosen using the vanilla model. Even if we could improve our score by choosing a different document, taking into account the pairwise scores with other permutation elements, this algorithm will not take advantage of it. Another way to look at this is, precision@1 would be no better than using the vanilla model of eq. (1).

The greedy procedure also permits beam search variants. Using a beam of $M$ candidates this gives a complexity of $O(Mk^2|\mathcal{D}|)$. This is tractable at test time, but the problem is that during (online) learning one would have to run this algorithm per query, which we believe is still too slow for the cases we consider here.

- Iterative search: Motivated by the defects in greedy search and LP relaxation, we propose one last, iterative method. This method is analogous to inference by iterated conditional modes in graphical models (Besag, 1986). (i) On iteration $t = 0$ predict with an unstructured model (i.e. do not use the second term involving $S$):

$$f_{iter:t=0}(q, \bar{d}) = \sum_{i=1}^{|\bar{d}|} w_i (qU^\top V \bar{d}_i). \quad (7)$$

As mentioned before, computing the best ranking $\bar{d}$ just involves sorting the scores $qU^\top V d_i$ and ordering the documents, largest first. Utilizing the sparse choice of $w_i = 1/i$, if $i \leq k$, and 0 otherwise described in Section 2.3 we do not have to sort the entire set, but are only required to find the top $k$ which can be done in $O(|\mathcal{D}|\log k)$ time using a heap. Let us denote the predicted ranked list as $\bar{d}^0$ and in general on each iteration $t$ we are going to make predictions $\bar{d}^t$. (ii) On subsequent iterations, we maximize the following scoring function:

$$f_{iter:t>0}(q, \bar{d}) = \sum_{i=1}^{|\bar{d}|} w_i (qU^\top V \bar{d}_i)$$

$$+ \sum_{i,j=1}^{|\bar{d}|} w_i w_j (\bar{d}_i^\top S^\top S \bar{d}_j^{t-1}). \quad (8)$$

As $\bar{d}^{t-1}$ is now fixed on iteration $t$, the per-document $d_i$ scores

$$(qU^\top V \bar{d}_i) + \sum_{j=1}^{|\bar{d}|} w_j (\bar{d}_i^\top S^\top S \bar{d}_j^{t-1}) \quad (9)$$

are now independent of each other. Hence, they can be calculated individually and, as before, can be sorted or the top $k$ can be found, dependent on the choice of $w$. If we use the sparse $w$ of eq. (4) (which we recommend) then the per-document scores are also faster to compute as we only require:

$$(qU^\top V \bar{d}_i) + \sum_{j=1}^{k} w_j (\bar{d}_i^\top S^\top S \bar{d}_j^{t-1}).$$

Overall this procedure then has complexity $O(Tk|\mathcal{D}|)$ when running for $T$ steps. While this does not look at first glance to be any faster than the greedy or beam search methods at testing time, it has important advantages at training time as we will see in the next section.

## 2.5 LEARNING

We are interested in learning a *ranking* function where the top $k$ retrieved items are of particular interest as they will be presented to the user. We wish to optimize all the parameters of our model jointly for that goal. As the datasets we intend to target are large scale, stochastic gradient descent (SGD) training seems a viable option. However, during training we cannot afford to perform full inference during each update step as otherwise training will be too slow. A standard loss function that already addresses that issue for the unstructured case which is often used for retrieval is the margin ranking criterion (Herbrich *et al.*, 2000; Joachims, 2002). In particular, it was also used for learning factorized document retrieval models in Bai *et al.* (2009). The loss can be written as:

$$err_{AUC} = \sum_{i=1}^{m} \sum_{d^- \neq d_i} \max(0, 1 - f(q_i, d_i) + f(q_i, d^-)). \quad (10)$$

For each training example $i = 1, \ldots, m$, the positive item $d_i$ is compared to all possible negative items $d^- \neq d_i$, and one assigns to each pair a cost if the negative item is larger or within a "margin" of 1 from the positive item. These costs are called *pairwise violations*. Note that all pairwise violations are considered equally if they have the same margin violation, independent of their position in the list. For this reason the margin ranking loss might not optimize the top $k$ very accurately as it cares about the average rank.

For the standard (unstructured) latent model case, the problem of optimizing the top of the rank list has also recently been addressed using sampling techniques (Weston *et al.*, 2011) in the so-called WARP (Weighted Approximately Ranked Pairwise) loss. Let us first write the predictions of our model for all items in the database as a vector $\bar{f}(q)$ where the $i^{th}$ element is $\bar{f}_i(q) = f(q, d_i)$. One then considers a class of ranking error functions:

$$err_{WARP} = \sum_{i=1}^{m} L(rank_{d_i}(\bar{f}(q_i))) \quad (11)$$

where $rank_{d_i}(\bar{f}(q_i))$ is the margin-based rank of the labeled item given in the $i^{th}$ training example:

$$rank_{d_i}(\bar{f}(q)) = \sum_{j \neq i} \theta(1 + \bar{f}_j(q) \geq \bar{f}_i(q)) \quad (12)$$

where $\theta$ is the indicator function, and $L(\cdot)$ transforms this rank into a loss:

$$L(r) = \sum_{i=1}^{r} \alpha_i, \text{ with } \alpha_1 \geq \alpha_2 \geq \cdots \geq 0. \quad (13)$$

The main idea here is to *weight* the pairwise violations depending on their position in the ranked list. Different choices of $\alpha$ define different weights (importance) of the relative position of the positive examples in the ranked list. In particular it was shown that by choosing $\alpha_i = 1/i$ a smooth weighting over positions is given, where most weight is given to the top position, with rapidly decaying weight for lower positions. This is useful when one wants to optimize precision at $k$ for a variety of different values of $k$ at once Usunier *et al.* (2009). (Note that choosing $\alpha_i = 1$ for all $i$ we have the same AUC optimization as equation (10)).

We can optimize this function by SGD following the authors of Weston *et al.* (2011), that is samples are drawn at random, and a gradient step is made for each draw. Due to the cost of computing the exact rank in (11) it is approximated by sampling. That is, for a given positive label, one draws negative labels until a violating pair is found, and then approximates the rank with

$$rank_d(\bar{f}(q)) \approx \left\lfloor \frac{|\mathcal{D}|-1}{N} \right\rfloor$$

where $\lfloor . \rfloor$ is the floor function, $|\mathcal{D}|$ is the number of items in the database and $N$ is the number of trials in the sampling step. Intuitively, if we need to sample more negative items before we find a violator then the rank of the true item is likely to be small (it is likely to be at the top of the list, as few negatives are above it).

This procedure for optimizing the top of the ranked list is very efficient, but it has a disadvantage with respect to structured learning: we cannot simply sample and score items any longer as we need to somehow score entire permutations. In particular, it is not directly applicable to several of the structured prediction approaches like LP, greedy or beam search. That is because we cannot compute the score of $\bar{f}_i$ independently because they depend on the ranking of all documents, which then makes the sampling scheme invalid. However, for (a variant of) the iterative algorithm which we described in the previous section the WARP (or AUC) technique can still be used.

The method is as follows. In the first iteration the model scores in eq. (7) are independent and so we can train using the WARP (or AUC) loss. We then have to compute $\bar{d}^0$ (the ranking of items) for each training example for use in the next iteration. Note that using the sparse $w$ of eq. (4) this is $O(\mathcal{D} \log k)$ to compute, and storage is also only a $|\mathcal{D}| \times k$ matrix of top items. After computing $\bar{d}^0$, in the second iteration we are again left with independent scoring functions $\bar{f}_i$ as long as we make one final modification, instead

**Algorithm 1** LASR training algorithm

**Input:** Training pairs $\{(q_i, d_i)\}_{i=1,\ldots,l}$.
Initialize model parameters $U_t, V_t$ and $S_t$ (we use mean 0, standard deviation $\frac{1}{\sqrt{d}}$) for each $t$.
**for** $t = 0, \ldots, T$ **do**
  **repeat**
    **if** $t = 0$ **then**
      $f(q, d) = q U_0^\top V_0 d$.
    **else**
      $f(q, d) = q U_t^\top V_t d + \sum_{j=1}^k w_j d^\top S_t^\top S_t \bar{d}_j^{t-1}$
    **end if**
    Pick a random training pair $(q, d^+)$.
    Compute $f(q, d^+)$.
    Set $N = 0$.
    **repeat**
      Pick a random document $d^- \in \mathcal{D}, d \neq d_i$.
      Compute $f(q, d^-)$.
      $N = N + 1$.
    **until** $f(q^+, d^+) < f(q^+, d^-) + 1$ or $N \geq |\mathcal{D}| - 1$
    **if** $f(q^+, d^+) < f(q^+, d^-) + 1$ **then**
      Make a gradient step to minimize:
      $L(\lfloor \frac{|\mathcal{D}|-1}{N} \rfloor) \max(1 - f(q^+, d^+) + f(q^+, d^-), 0)$.
      Project weights to enforce constraints,
      i.e. if $||U_{ti}|| > C$ then $U_{ti} \leftarrow (CU_{ti})/||U_{ti}||$,
      $i = 1, \ldots, D$ (and likewise for $V_t$ and $S_t$).
    **end if**
  **until** validation error does not improve.
  For each training example, compute the top $k$ ranking documents $\bar{d}_i^t$, $i = 1, \ldots, k$ for iteration $t$ using $f(q, d)$ defined above.
**end for**

of using eq. (8) we instead use:

$$f_{iter:t>0}(q, \bar{d}) = \sum_{i=1}^{|\bar{d}|} w_i (q U_t^\top V_t \bar{d}_i)$$
$$+ \sum_{i,j=1}^{|\bar{d}|} w_i w_j (\bar{d}_i^\top S_t^\top S_t \bar{d}_j^{t-1}). \quad (14)$$

on iteration $t$, where $U_t$, $V_t$ and $S_t$ are separate matrices for each iteration. This decouples the learning at each iteration. Essentially, we are using a cascade-like architecture of $t$ models trained one after the other. Note that if a global optimum is reached for each $t$ then the solution should always be the same or improve over step $t - 1$, as one could pick the weights that give exactly the same solution as for step $t - 1$.

So far, the one thing we have failed to mention is regularization during learning. One can regularize the parameters by preferring smaller weights. We constrain them using $||S_{ti}|| \leq C$, $||U_{ti}|| \leq C$, $||V_{ti}|| \leq C$, $i = 1, \ldots, |\mathcal{D}|$. During SGD one projects the parameters back on to the constraints at each step, following the same procedure used in several other works, e.g. Weston *et al.* (2011); Bai *et al.* (2009). We can optimize hyperparameters of the model such as $C$ and the learning rate for SGD using a validation set.

Overall, our preferred version of Latent Structured Ranking that combines all these design decisions is given in Algorithm 1.

## 3 PRIOR WORK

In the introduction we already mentioned several latent ranking methods: SVD (Billsus and Pazzani, 1998; Bell *et al.*, 2009), NMF (Lee and Seung, 2001), LSI (Deerwester *et al.*, 1990), PSI (Bai *et al.*, 2009) and WSABIE (Weston *et al.*, 2011). We should mention that many other methods exist as well, in particular probabilistic methods like pLSA (Hofmann, 1999) and LDA (Blei *et al.*, 2003). None of those methods, whether they are supervised or unsupervised, take into the structure of the ranked list as we do in this work, and we will use several of them as baselines in our experiments.

There has been a great deal of recent work on structured output learning (Bakir *et al.*, 2007), particularly for linear or kernel SVMs (which are not latent embedding methods). In methods like Conditional Random Fields (Lafferty *et al.*, 2001), SVM-struct (Tsochantaridis *et al.*, 2004) LaSO (Daumé III and Marcu, 2005) and SEARN (Daumé *et al.*, 2009) one learns to predict an output which has structure, e.g. for sequence labeling, parse tree prediction and so on. Predicting ranked lists can also be seen in this framework. In particular LaSO (Daumé III and Marcu, 2005) is a general approach that considers approximate inference using methods like greedy approximation or beam search that we mentioned in Section 2.4. As we said before, due to the large number of items we are ranking many of those approaches are infeasible. In our method, scalabality is achieved using a cascade-like training setup, and in this regard is related to (Weiss *et al.*, 2010). However, unlike that work, we do not use it to prune the set of items considered for ranking, we use it to consider pairwise item similarities.

The problem of scoring entire permutations for ranking is well-known and has been investigated by many authors (Yue *et al.*, 2007a,b; Le and Smola, 2007; Xia *et al.*, 2008). These works have primarily focused on on using knowledge of the structure (in this case the predicted positions in the ranked list) in order to optimize the right metric, e.g. MAP or precision@k. In that sense, methods like WSABIE which uses the WARP loss already use structure in the same way. In our work we also optimize top-of-the-ranked-list metrics by using WARP, but in addition we also use the ranking structure to make predictions dependent on the query and the other predicted items during inference by encoding this in the model itself. That is, in our work we explicitly seek to use (and learn) inter-document similarity measures.

There has been work on taking into account inter-document similarities during ranking. The most famous and prominent idea is pseudo-relevance feedback via query expansion (Rocchio, 1971). Pseudo-relevance works by doing normal retrieval (e.g. using cosine similarity in a vector space model), to find an initial set of most relevant documents, and then assuming that the top $k$ ranked documents are relevant, and performing retrieval again by adjusting the cosine similarity based on previously retrieved documents. In a sense, LaSR is also a pseudo-relevance feedback technique, but where inter-document similarities are learned to minimize ranking loss.

More recently, some authors have investigated incorporating inter-document similarity during ranking. Qin *et al.* (2008) have investigated incorporating a fixed document-document similarity feature in ranking. In their work, however, they did not score permutations. Instead, each document was associated with a relevance score and the authors treated learning as a structured regression problem. For a situation with implicit feedback, Raman *et al.* (2012) investigate an inference technique similar to our greedy algorithm. Volkovs and Zemel (2009) also explored listwise ranking using pairwise document interactions in a probabilistic setup. To the best of our knowledge, however, none of these methods investigate a learned inter-document similarity (i.e. latent parameters for that goal), which is the most powerful feature of LaSR.

## 4 EXPERIMENTS

We considered two large scale tasks to test our proposed method. The first is a music recommendation task with over 170,000 artists (possible queries or items) and 5 million training pairs. The second is a task of large scale image annotation with over 15,000 labels and 7 million training examples.

### 4.1 MUSIC RECOMMENDATION TASK

The first task we conducted experiments on is a large scale music recommendation task. Given a query (seed) artist, one has to recommend to the user other artists that go well together with this artist if one were listening to both in succession, which is the main step in playlisting and artist page recommendation on sites like `last.fm`, `music.google.com` and programs such as iTunes and `http://the.echonest.com/`.

Table 1: Recommendation Results on the music recommendation task. We report for recall at 5, 10, 30 and 50 for our method and several baselines.

| Method | | R@5 | R@10 | R@30 | R@50 |
|---|---|---|---|---|---|
| NMF | | 3.76% | 6.38% | 13.3% | 17.8% |
| SVD | | 4.01% | 6.93% | 13.9% | 18.5% |
| LaSR | ($t=0$) | 5.60% | 9.49% | 18.9% | 24.8% |
| LaSR | ($t=1$) | 6.65% | 10.73% | 20.1% | 26.7% |
| LaSR | ($t=2$) | 6.93% | 10.95% | 20.3% | 26.5% |

Table 2: Changing the embedding size on the music recommendation task. We report R@5 for various dimensions $n$.

| Method | $n=$ | 25 | 50 | 100 | 200 |
|---|---|---|---|---|---|
| NMF | | 2.82% | 3.76% | 3.57% | 4.82% |
| SVD | | 3.61% | 4.01% | 4.53% | 5.28% |
| LaSR | ($t=0$) | 5.23% | 5.60% | 6.24% | 6.42% |

We used the "Last.fm Dataset - 1K users" dataset available from http://www.dtic.upf.edu/~ocelma/MusicRecommendationDataset/lastfm-1K.html. This dataset contains (user, timestamp, artist, song) tuples collected from the Last.fm (www.lastfm.com) API, representing the listening history (until May 5th, 2009) for 992 users and 176,948 artists. Two consecutively played artists by the same user are considered as a (query, item) pair. Hence, both $q_i$ and $d_i$ are $D = 176,948$ sparse vectors with one non-zero value (a one) indicating which artist they are. One in every five days (so that the data is disjoint) were left aside for testing, and the remaining data was used for training and validation. Overall this gave 5,408,975 training pairs, 500,000 validation pairs (for hyperparameter tuning) and 1,434,568 test pairs.

We compare our Latent Structured Ranking approach to the same approach *without* structure by only performing one iteration of Algorithm 1. We used $k = 20$ for eq. (4). We also compare to two standard methods of providing latent recommendations, Singular Value Decomposition (SVD) and Non-negative Matrix Factorization (NMF). For SVD the Matlab implementation is used, and for NMF the implementation at http://www.csie.ntu.edu.tw/~cjlin/nmf/ is used.

**Main Results** We report results comparing NMF, SVD and our method, Latent Structured Ranking (LaSR) in Table 1. For every test set (query, item) pair we rank the document set $\mathcal{D}$ according to the query and record the position of the item in the ranked list. We then measure the recall at 5, 10, 30 and 50. (Note that because there is only one item in the pair, precision at k is equal to recall@k divided by k.) We then average the results over all pairs. For all methods the latent dimension $n = 50$. For LaSR, we give results for iterations $t = 0, \ldots, 2$, where $t = 0$ does not use the structure. LaSR with $t = 0$ already outperforms SVD and NMF. LaSR optimizes the top of the ranked list at training time (via the WARP loss), whereas SVD and NMF do not, which explains why it can perform better here on top-of-the-list metrics. We tested LaSR $t = 0$ using the AUC loss (10) instead of WARP (11) to check this hypothesis and we obtained a recall at 5, 10, 30 and 50 of 3.56%, 6.32%, 14.8% and 20.3% respectively which are slightly worse, than, but similar to SVD, thus confirming our hypothesis. For LaSR with $t = 1$ and $t = 2$ our method takes into account the structure of the ranked list at inference time, $t = 1$ outperforms iteration $t = 0$ that does not use the structure. Further slight gains are obtained with another iteration ($t = 2$).

**Changing the Embedding Dimension** The results so far were all with latent dimension $n = 50$. It could be argued that LaSR with $t > 0$ has more capacity (more parameters) than competing methods, and those methods could have more capacity by increasing their dimension $n$. We therefore report results for various embedding sizes ($n = 10, 25, 50, 100$) in Table 2. The results show that LaSR ($t = 0$) consistently outperforms SVD and NMF for all the dimensions tried, but even with 200 dimensions, the methods that do not use structure (SVD, NMF and LaSR $t = 0$) are still outperformed by LaSR that does use structure ($t > 0$) even with $n = 50$ dimensions.

**Analysis of Predictions** We give two example queries and the top ranked results for LaSR with and without use of structure (($t = 0$) and ($t = 1$)) in Table 3. The left-hand query is a popular artist "Bob Dylan". LaSR ($t = 0$) performs worse than ($t = 1$) with "Wilco" in positon 1 - the pair ("Bob Dylan","Wilco") only appears 10 times in the test set, whereas ("Bob Dylan", "The Beatles") appears 40 times, and LaSR ($t = 1$) puts the latter in the top position. In general $t = 1$ improves the top ranked items over $t = 0$, removing or demoting weak choices, and promoting some better choices. For example, "Sonic Youth" which is a poor match is demoted out of the top 20. The second query is a less popular artist "Plaid" who make electronic music. Adding structure to LaSR again improves the results in this case by boosting relatively more popular bands like "Orbital" and "$\mu - ziq$", and relatively more related bands like "Four-Tet" and "Squarepusher", whilst demoting some lesser known bands.

Table 3: Music Recommendation results for our method LaSR with ($t = 1$) and without ($t = 0$) using the structure. We show top ranked results for a popular query, "Bob Dylan" (folk rock music) and a less popular query "Plaid" (electronic music). Total numbers of train and test pairs for given artist pairs are in square brackets, and totals for all artists shown are given in the last row. Artists where the two methods differ are labeled with an asterisk. Adding structure improves the results, e.g. unrelated bands like "Sonic Youth" are demoted in the "Bob Dylan" query, and relatively more popular bands like Orbital and $\mu-$ziq and more related bands like Four-Tet and Squarepusher are boosted for the "Plaid" query.

| LaSR $t=0$ (no structure) | LaSR $t=1$ (structured ranking) | LaSR $t=0$ (no structure) | LaSR $t=1$ (with structured ranking) |
|---|---|---|---|
| QUERY: BOB DYLAN | QUERY: BOB DYLAN | QUERY: PLAID | QUERY: PLAID |
| WILCO [53,10] | THE BEATLES [179,40] | BOARDS OF CANADA [27,5] | BOARDS OF CANADA [27,5] |
| THE ROLLING STONES [40,9] | RADIOHEAD [61,16] | AUTECHRE [13,1] | APHEX TWIN [9,3] |
| THE BEATLES [179,40] | THE ROLLING STONES [40,9] | APHEX TWIN [9,3] | AUTECHRE [13,1] |
| R.E.M. [35,12] | JOHNNY CASH [49,11] | BIOSPHERE [6,1] | BIOSPHERE [6,1] |
| JOHNNY CASH [49,11] | THE CURE [38,10] | WAGON CHRIST [3,1] | SQUAREPUSHER [11,2] |
| BECK [42,18] | DAVID BOWIE [48,12] | AMON TOBIN [6,3] | FUTURE SOUND OF LONDON [5,2] |
| DAVID BOWIE [48,12] | WILCO [53,10] | AROVANE [5,1] | FOUR TET* [6,2] |
| PIXIES [31,4] | PINK FLOYD* [30,8] | FUTURE SOUND OF LONDON [5,2] | MASSIVE ATTACK* [4,2] |
| BELLE AND SEBASTIAN [28,4] | U2* [40,16] | THE ORB [3,2] | AROVANE [5,1] |
| THE BEACH BOYS* [22,6] | THE SMITHS [25,7] | SQUAREPUSHER [11,2] | AIR [9] |
| THE CURE [38,10] | SUFJAN STEVENS [23,4] | BOLA [5,2] | THE ORB [3,2] |
| ARCADE FIRE* [34,6] | R.E.M. [35,12] | CHRIS CLARK* [4,2] | ISAN [3,1] |
| RADIOHEAD [61,16] | BELLE AND SEBASTIAN [28,4] | KETTEL* [3,0] | AMON TOBIN [6,3] |
| SONIC YOUTH* [35,9] | BECK [42,18] | ULRICH SCHNAUSS [7,1] | BOLA [5,2] |
| BRUCE SPRINGSTEEN [41,11] | THE SHINS* [22,13] | APPARAT* [3,0] | ORBITAL* [7,1] |
| THE SMITHS [25,7] | PIXIES [31,4] | ISAN [3,1] | MURCOF [4] |
| THE VELVET UNDERGROUND* [29,11] | RAMONES* [36,8] | AIR [9] | ULRICH SCHNAUSS [7,1] |
| SUFJAN STEVENS [23,4] | BRUCE SPRINGSTEEN [44,11] | CLARK* [0,0] | WAGON CHRIST [3,1] |
| TOM WAITS* [19,13] | DEATH CAB FOR CUTIE* [32,7] | MURCOF [4,0] | $\mu$-ZIQ* [5,3] |
| (TRAIN,TEST) TOTALS = [835,213] | (TRAIN,TEST) TOTALS = [856,220] | (TRAIN,TEST) TOTALS = [126,27] | (TRAIN,TEST) TOTALS = [138,33] |

Table 4: Image Annotation Results comparing WSABIE with LaSR. The top 10 labels of each method is shown, the correct label (if predicted) is shown in bold. In many cases LaSR can be seen to improve on WSABIE by demoting bad predictions e.g. war paint, soccer ball, segway, denture, rottweiler, reindeer, tv-antenna, leopard frog (one example from each of the first 8 images). In the last 3 images neither method predicts the right label (armrest, night snake and heifer) but LaSR seems slightly better (e.g. more cat, snake and cow predictions).

| INPUT IMAGE | WSABIE | LaSR | INPUT IMAGE | WSABIE | LaSR |
|---|---|---|---|---|---|
| | WORKROOM, LIFE OFFICE, WAR PAINT, **day nursery**, HOMEROOM, SALON, FOUNDLING HOSPITAL, TEACHER, SEWING ROOM, CANTEEN | WORKROOM, SALON, **day nursery**, SCHOOLROOM, STUDENT, HOMEROOM, CLOTHESPRESS, DAY SCHOOL, SEWING ROOM, STUDY | | TV-ANTENNA, TRANSMISSION LINE, SHEARS, REFRACTING TELESCOPE, SCISSORS, ELECTRICAL CABLE, CHAIN WRENCH, ASTRONOMICAL TELESCOPE, WIRE, BOAT HOOK, | SCISSORS, SHEARS, TINSNIPS, WINDMILL, GARDEN RAKE, TV-ANTENNA, FORCEPS, SAFETY HARNESS, **medical instrument**, PLYERS |
| | SOCCER BALL, TENT, INFLATED BALL, FLY TENT, SHELTER TENT, CAMPING, **pack tent**, MAGPIE, FIELD TENT, WHITE ADMIRAL BUTTERFLY | SHELTER TENT, TENT, CAMPING, **pack tent**, FLY TENT, MOUNTAIN TENT, TENT FLAP, TWO-MAN TENT, FIELD TENT, POP TENT | | LEOPARD FROG, SAUCEBOAT, SUGAR BOWL, CREAM PITCHER, TUREEN, SPITTOON, PITCHER, EARTHENWARE, RAINBOW FISH, SLOP BOWL | TUREEN, GLAZED EARTHENWARE, SLOP BOWL, SAUCEBOAT, SPITTOON, **cullender**, EARTHENWARE, PUNCH BOWL, SUGAR BOWL, CREAM PITCHER |
| | ROLLER SKATING, SKATEBOARD, CROSS-COUNTRY SKIING, SKATING, SEGWAY, HOCKEY STICK, SKATEBOARDING, SKATING RINK, CRUTCH, ROLLER SKATE WHEEL | ROLLER SKATE WHEEL, ROLLER SKATING, SKATEBOARD, IN-LINE SKATE, **roller skate**, CROSS-COUNTRY SKIING, SKATEBOARDING, UNICYCLE, SKATE, SKATING | | REDPOLL, FROGMOUTH, SCREECH OWL, POSSUM, GREY PARROT, FINCH, GYPSY MOTH, GRAY SQUIRREL, LYCAENID BUTTERFLY, CAIRN TERRIER | FROGMOUTH, SOFT-COATED WHEATEN TERRIER, GRAY SQUIRREL, CAIRN TERRIER, GYPSY MOTH, TABBY CAT, CHINCHILLA LANIGER, EGYPTIAN CAT, BURMESE CAT, ABYSSINIAN CAT |
| | PARTIAL DENTURE, ROUND-BOTTOM FLASK, **flask**, PANTY GIRDLE, NIGHT-LIGHT, PATCHOULY, ORGANZA, ORGANDY, BABY, WIG | ROUND-BOTTOM FLASK, **flask**, RUMMER, TORNADO LANTERN, SPOTLIGHT, FOETUS, OIL LAMP, SCONCE, INFRARED LAMP, DECANTER | | GRASS SNAKE, GARTER SNAKE, RIBBON SNAKE, COMMON KINGSNAKE, BLACK RATTLER, WESTERN RIBBON SNAKE, EUROPEAN VIPER, PICKEREL FROG SPINY ANTEATER, EASTERN GROUND SNAKE | GARTER SNAKE, EUROPEAN VIPER, GRASS SNAKE, CALIFORNIA WHIPSNAKE, COMMON KINGSNAKE, EASTERN GROUND SNAKE, NORTHERN RIBBON SNAKE, PUFF ADDER, WHIPSNAKE, BLACK RATTLER |
| | RUBBER BOOT, ROTTWEILER, PILLAR, **combat boot**, CALF, RIDING BOOT, TROUSER, WATCHDOG, SHEPHERD DOG, LEG | **combat boot**, RIDING BOOT, RUBBER BOOT, LEG COVERING, RUBBER, TROUSER, TABIS BOOT, TROUSER LEG, BOOT, LAWN TOOL | | | |
| | REAPER, REINDEER, CANNON, STEAMROLLER, SEEDER, PLOW, TRACTOR, COMBINE, CANNON, CARIBOU | REAPER, SEEDER, **gun carriage**, FARM MACHINE, COMBINE, PLOW, CARIBOU, HAYMAKER, TRACTOR, TRENCH MORTAR, | | BLACK BEAR, BLACK VULTURE, LABIATED BEAR, GREATER GIBBON, VULTURE, AMERICAN BLACK BEAR, CUCKOO, BLACK SHEEP, BUFFALO, YAK, | BLACK BEAR, YAK, BEAR, AMERICAN BLACK BEAR, STOCKY HORSE, CALF, WILD BOAR, BULL, DRAFT HORSE, BLACK ANGUS |

Table 5: **Summary of Test Set Results on Imagenet.** Recall at 1, 5, 10, Mean Average Precision and Mean Rank are given.

| Algorithm | r@1 | r@5 | r@10 | MAP | MR |
|---|---|---|---|---|---|
| One-vs-Rest | 2.83% | 8.48% | 13.2% | 0.065 | 667 |
| Rank SVM | 5.35% | 14.1% | 19.3% | 0.102 | 804 |
| WSABIE | 8.39% | 19.6% | 26.3% | 0.144 | 626 |
| LaSR ($t=1$) | 9.45% | 22.1% | 29.1% | 0.161 | 523 |

## 4.2 IMAGE ANNOTATION TASK

ImageNet (Deng *et al.*, 2009) (http://www.image-net.org/) is a large scale image database organized according to WordNet (Fellbaum, 1998). WordNet is a graph of linguistic terms, where each concept node consists of a word or word phrase, and the concepts are organized within a hierarchical structure. ImageNet is a growing image dataset that attaches quality-controlled human-verified images to these concepts by collecting images from web search engines and then employing annotators to verify whether the images are good matches for those concepts, and discarding them if they are not. For many nouns, hundreds or even thousands of images are labeled. We can use this dataset to test image annotation algorithms. We split the data into train and test and try to learn to predict the label (annotation) given the image. For our experiments, we downloaded the "Spring 2010" release which consists of 9 million images and 15,589 possible concepts (this is a different set to (Weston *et al.*, 2011) but our baseline results largely agree). We split the data into 80% for training, 10% for validation and 10% for testing.

Following (Weston *et al.*, 2011) we employ a feature representation of the images which is an ensemble of several representations which is known to perform better than any single representation within the set (see e.g. (Makadia *et al.*, 2008)). We thus combined multiple feature representations which are the concatenation of various spatial (Grauman and Darrell, 2007) and multiscale color and texton histograms (Leung and Malik, 1999) for a total of about $5 \times 10^5$ dimensions. The descriptors are somewhat sparse, with about 50,000 non-zero weights per image. Some of the constituent histograms are normalized and some are not. We then perform Kernel PCA (Schoelkopf *et al.*, 1999) on the combined feature representation using the intersection kernel (Barla *et al.*, 2003) to produce a 1024 dimensional input vector for training.

We compare our proposed approach to several baselines: one-versus-rest large margin classifiers (One-vs-Rest) of the form $f_i(x) = w_i^\top x$ trained online to perform classification over the 15,589 classes, or the same models trained with a ranking loss instead, which we refer to as RANK SVM, as it is an online version of the pairwise multiclass (ranking) loss of (Weston and Watkins, 1999; Crammer and Singer, 2002). Finally, we compare to WSABIE a (unstructured) latent ranking method which has yielded state-of-the-art performance on this task. For all methods, hyperparameters are chosen via the validation set.

**Results** The overall results are given in Table 5. One-vs-Rest performs relatively poorly (2.83% recall@1), perhaps because there are so many classes (over 15,000) that the classifiers are not well calibrated to each other (as they are trained independently). The multiclass ranking loss of Rank SVM performs much better (5.35% recall@1) but is still outperformed by WSABIE (8.39% recall@1). WSABIE uses the WARP loss to optimize the top of the ranked list and its good performance can be explained by the suitability of this loss function for measure like recall@k. LaSR with $t=0$ is essentially identical to WSABIE in this case and so we use that model as our "base learner" for iteration 0. LaSR ($t=1$), that does use structure, outperforms WSABIE with a recall@1 of 9.45%.

Some example annotations are given in Table 4. LaSR seems to provide more consistent results than WSABIE on several queries (with less bad predictions in the top $k$) which improves the overall results, whilst maintaining the right level of diversity on others.

## 5 CONCLUSION

In this paper we introduced a method for learning a latent variable model that takes into account the structure of the predicted ranked list of items given the query. The approach is quite general and can potentially be applied to recommendation, annotation, classification and information retrieval tasks. These problems often involve millions of examples or more, both in terms of the number of training pairs and the number of items to be ranked. Hence, many otherwise straight-forward approaches to structured prediction approaches might not be applicable in these cases. The method we proposed is scalable to these tasks.

Future work could apply latent structured ranking to more applications, for example in text document retrieval. Moreover, it would be interesting to explore using other algorithms as the "base algorithm" which we add the structured predictions to. In this work, we used the approach of (Weston *et al.*, 2011) as our base algorithm, but it might also be possible to make structured ranking versions of algorithms like Non-negative matrix factorization, Latent Semantic Indexing or Singular Value Decomposition as well.